\documentclass{article}
\usepackage{spconf,amsmath,graphicx}
\usepackage{xcolor}
\usepackage{booktabs}
\usepackage{multirow}
\usepackage{siunitx}
\usepackage{hyperref}

\title{Hand-Based Person Identification using Global and Part-Aware Deep Feature Representation Learning}
\name{
\begin{tabular}{c} Nathanael L. Baisa, Bryan Williams, Hossein Rahmani, Plamen Angelov, Sue Black \end{tabular} 
\thanks{
Nathanael L. Baisa is with the School of Computer Science and Informatics, De Montfort University, Leicester LE1 9BH, UK. Email: nathanael.baisa@dmu.ac.uk. }
\thanks{
Bryan Williams, Hossein Rahmani, Plamen Angelov and Sue Black are with the School of Computing and Communications, Lancaster University, Lancaster LA1  4WA, UK. Emails: \{b.williams6, h.rahmani, p.angelov, sue.black\}@lancaster.ac.uk. }
}
\address{}

\begin{document}
%
\maketitle
\begin{abstract}

In cases of serious crime, including sexual abuse, often the only available information with demonstrated potential for identification is images of the hands. Since this evidence is captured in uncontrolled situations, it is difficult to analyse. As global approaches to feature comparison are limited in this case, it is important to extend to consider local information. In this work, we propose hand-based person identification by learning both global and local deep feature representations. Our proposed method, Global and Part-Aware Network (GPA-Net), creates global and local branches on the conv-layer for learning robust discriminative global and part-level features. For learning the local (part-level) features, we perform uniform partitioning on the conv-layer in both horizontal and vertical directions. We retrieve the parts by conducting a soft partition without explicitly partitioning the images or requiring external cues such as pose estimation. We make extensive evaluations on two large multi-ethnic and publicly available hand datasets, demonstrating that our proposed method significantly outperforms competing approaches. 

\end{abstract}

\section{Introduction} \label{sec:intro}

There are many human parts or traits that can help to uniquely identify an individual such as face~\cite{DenGuoZaf19, ZhaZhuShi17}, body~\cite{YeSheLin2020,Nat21}, hand~\cite{Mah19,AttAkhCha21}, fingerprint~\cite{EngCaoJai19}, Iris~\cite{ZhaKum17}, gait~\cite{MurShiMak15} and voice~\cite{ChuNagZis18}. A survey of person identification based on different human traits is given in~\cite{DanEliRos16, WanWei20, YeSheLin2020}. These are primary biometric traits also referred to as biometric modalities. Personal attributes or ancillary information such as gender, age, (hair, skin, eye) color, ethnicity, height, weight, etc. which are also known as soft biometrics can be cheaply obtained from the primary biometric traits~\cite{DanEliRos16}. 
The identification of an individual can be improved by fusing different primary biometric traits such as face and body or by fusing biometric modalities with soft biometrics such as face and gender~\cite{SinSinRos19}, however, enough disriminative information may not be obtained by using only the soft biometric traits to uniquely identify a subject. 

Often, the only avaliable information relating to the identity of the perpetrator in cases of serious crime is images of the hand. Hand images, one of the primary biometric traits, provide distinctive features for person identification. They also have less variability when compared to other biometric modalities such as faces~\cite{DanEliRos16}. Facial expression has a considerable effect on the variability of the comparable features, whereas the hand has more limited flexibility, suggesting its potential as a more robust biometric trait for identification. Complex security systems rely on detailed images of the hand vein patterns with infrared imaging~\cite{SanRamCor11} but this is very rarely available in real cases. Evidence images of the hands are commonly captured with digital cameras~\cite{Mah19} and there is a strong need to investigate the potential for identification from digital images of the hand.

A method in~\cite{Mah19} trained a two-stream convolutional neural network (CNN) and used it as a feature extractor to obtain CNN-features which have been fed into a set of support vector machine (SVM) classifiers for a hand-based person identification, however, this method is not an end-to-end. In addition, this method treats images of a subject's right and left hands as the same which is not usual, due to the disparity in comparable features including vein patterns~\cite{WanCaoTan19}. The work in~\cite{YimChaShu20} used rather similar approach with additional data type for fusion, near-infrared (NIR) images. Researchers have also demonstrated that leveraging human part cues from external pose estimation to alleviate pose variations and learning feature representations can improve person re-identification based on images of the body~\cite{SuLiZha17}. However, the errors in the pose estimation can propagate to the identification which limits its performance. To allievate this problem, uniform partitioning for learning part-level features has been proposed in~\cite{YifLiaYi18}, however, this method considered only the part-level features, partitioned in only horizontal stripes, and has been designed for body-based person re-identification.


In this work, we propose a hand-based person identification method, Global and Part-Aware Network (GPA-Net), by learning both global and local deep feature representations from hand images captured by digital cameras. The proposed method follows an end-to-end training. Our contributions can be summarized as follows.
\begin{enumerate}
\item We extend the capability of uniform partitioning for hand-based person identification by including both horizontal and vertical partitioning.
\item We incorporate global representation learning in addition to the local (part-level) features proposing a combined approach that is flexible in terms of the backbone architecture for hand-based person recognition.
\item We make extensive evaluations on two large multi-ethnic and public hand datasets (11k hands~\cite{Mah19} and HD~\cite{KumXu16}  datasets).
\end{enumerate}

The rest of the paper is organized as follows. The proposed method including the overall architecture of GPA-Net is described in Section~\ref{sec:proposedMethod} followed the experimental results in Section~\ref{sec:experimentalResults}. The main conclusion along with suggestion for future work is summarized in Section~\ref{sec:Conclusion}.

\section{Proposed Method} \label{sec:proposedMethod}

\begin{figure*}[t]
\begin{center}
  \includegraphics[width=0.8\linewidth]{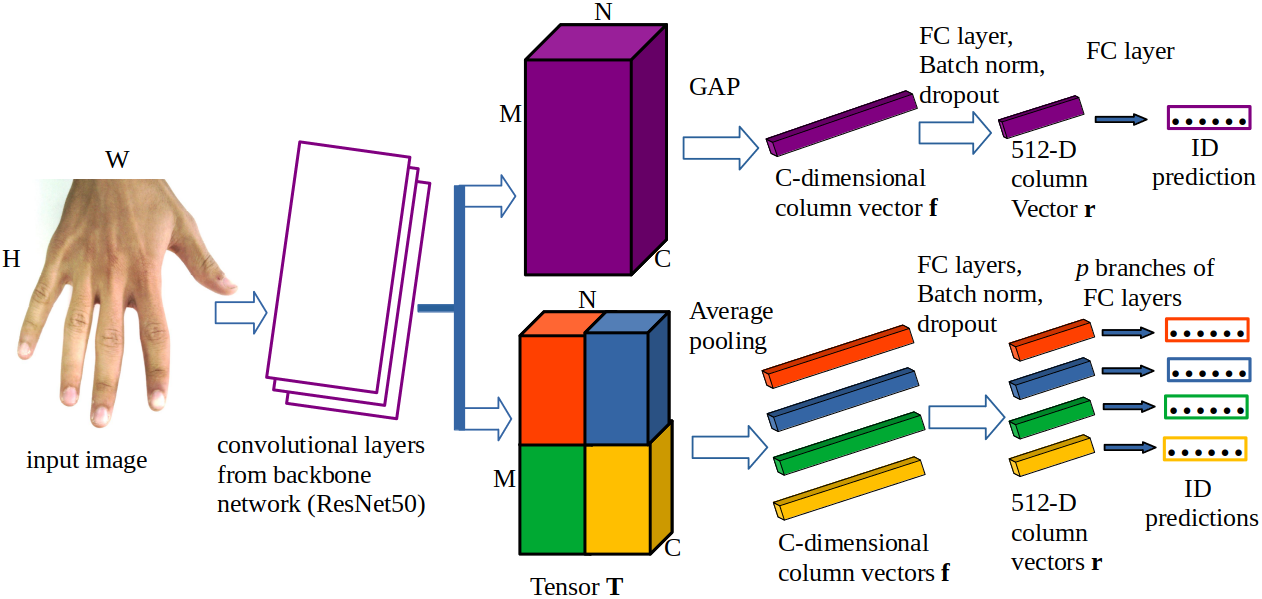} \\
\end{center}
   \caption{Structure of GPA-Net. Given an input image, two separate 3D tensors (one for global branch and the other for local branch) are obtained by passing it through the stacked convolutional layers from the backbone network. Each classifier of the local or the global branch predicts the identity of the input image during training.}
\label{fig:GPA-Net}
\end{figure*}

Our proposed method (GPA-Net) is summarized in Fig.~\ref{fig:GPA-Net}. We use ResNet50~\cite{HeZhaSun16} pretrained on ImageNet as a backbone network due to its concise architecture with competitive performance though any network designed for image classification, for instance Google Inception~\cite{ChrSerVin17}, can be adapted by removing the hidden fully-connected layers. In our case, we make some modifications to the ResNet50 architecture to produce the GPA-Net. The structure of the original ResNet50 before the global average pooling (GAP) layer remains the same. When an input image passes through the backbone network, it becomes a 3D tensor \textbf{T} of activations just before it enters the GAP layer. At this stage, we create two branches: Global and Local branches explained as follows.

\noindent \textbf{Global branch:} For this branch, the GAP layer is used to summarize the 3D tensor \textbf{T} of activations to form a C-dimensional (2048-dimensional in case of a ResNet50 backbone network) column feature vector $\textbf{f}_g$. C is the number of channels of the tensor (feature map) \textbf{T}.

\noindent \textbf{Local branch:} In this case, we change from the GAP layer to conventional average pooling layer to create uniform partitions on the 3D tensor \textbf{T} of activations along both horizontal and vertical directions to learn part-level features $\textbf{f}_l$ where $l \in [1, p]$; $p$ is the total number of partitions.

For both branches, new fully-connected layers (FCs) are added to reduce the C-dimensional column feature vectors \textbf{f} to 512-dimensional feature vectors \textbf{r}. We also include a batch normalization and dropout with probability of 0.5 to reduce possible over-fitting. Each dimensionally reduced feature vector \textbf{r} is given as input to a classification layer which is implemented using a FC layer followed by a softmax function. The classification layer predicts the identity (ID) of each input. Given the input image size ([H, W]), the spatial size of the tensor \textbf{T} ([M, N ]) is determined by the spatial down-sampling ratio of the backbone network. To increase the size of the tensor \textbf{T} which in turn increases the performance as observed in~\cite{YifLiaYi18,LuoGuLia19}, we remove the last spatial down-sampling operation by changing the last stride from 2 to 1 in the backbone network.

We optimize the GPA-Net during training by minimizing the sum of the cross-entropy losses over $p$ pieces of ID predictions of the Local branch and ID prediction of the Global branch i.e. each classifier, either of local or global branch, predicts the identity of the input image. The total cross-entropy loss is given as

\begin{equation}
    \mathcal{L}_{total} = \underbrace{\sum_{i=1}^K-q_{i,g} \log(p_{i,g})}_{Global~loss} + \lambda \underbrace{\sum_{l=1}^p \sum_{i=1}^K-q_{i,l} \log(p_{i,l})}_{Local~loss} 
\label{eqn:totalLoss}
\end{equation}
\noindent where $K$ is the total number of classes (identities). $p_{i,\gamma}$ is the predicted probability of class $i$; $\gamma = l$ in case of learning part-level features and $\gamma = g$ in case of learning global features. $\lambda$ is the weight balancing the losses of the two branches. $\lambda = 1$ is our standard choice; $\lambda = \frac{1}{p}$ decreases the performance, for instance. We also use label smoothing~\cite{SzeVanIof16} which prevents over-fitting and over-confidence with smoothing value ($\epsilon$) of 0.1. The ground-truth distribution over labels $q_{i,\gamma}$ by including label smoothing can be given as

\begin{equation}
    q_{i,\gamma} =
\begin{cases}
    1 - \frac{K-1}{K} \epsilon, & \text{if } i = y\\
    \frac{1}{K} \epsilon,              & \text{otherwise}
\end{cases}
\label{eq:labelSmoothing}
\end{equation}
\noindent where $y$ is ground-truth label. $\gamma$ is equal to $g$ for computing global loss or $l$ otherwise.

During testing, we concatenate all C-dimensional (2048-D in our case) feature vectors of $p$ pieces of Local branch and global branch i.e. $\mathcal{F} = [\textbf{f}_g, \textbf{f}_1, \textbf{f}_2, ..., \textbf{f}_p]$, and then compare features of each query image with gallery features using cosine distance. Concatenating the global and part-level features in this way has a stable performance on all datasets, for instance, when compared to taking the mean of the part-level features and then concatenating to the global features.



\noindent

\section{Experimental Results}  \label{sec:experimentalResults}

\subsection{Settings}

\begin{figure}[!h] 
\begin{minipage}[b]{1.0\linewidth}
  \centering
  \centerline{\includegraphics[width=8.5cm]{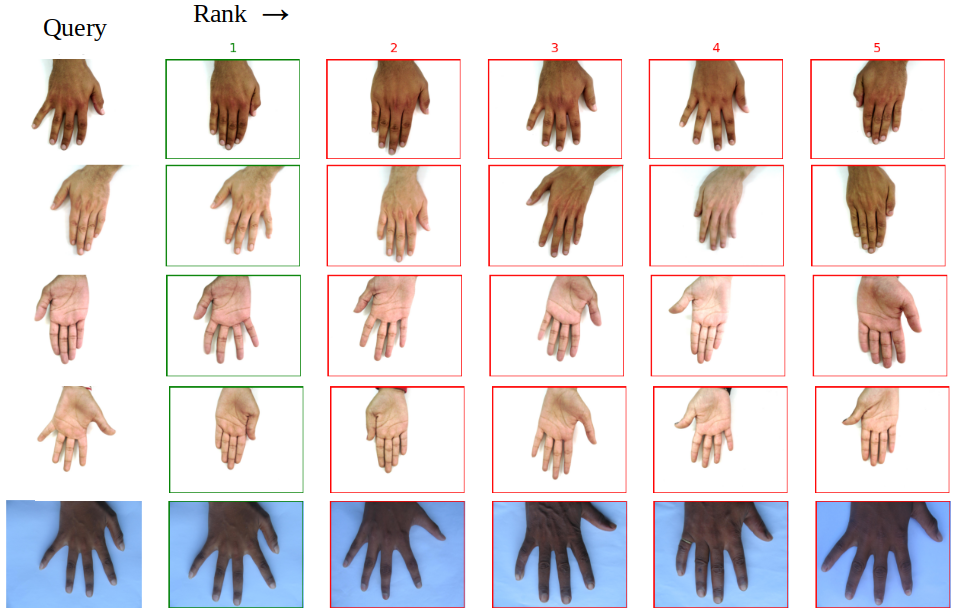}}
\end{minipage}
\caption{Some qualitative results of our method using query vs ranked results retrieved from gallery. From top to bottom row are right dorsal of 11k, left dorsal of 11k, right palmar of 11k, left palmar of 11k and HD datasets. The green and red bounding boxes denote the correct and the wrong matches, respectively.}
\label{fig:result}
\end{figure}

\textbf{Datasets:} We use two datasets for evaluation, 11k hands\footnote{\url{https://sites.google.com/view/11khands}} dataset~\cite{Mah19} and Hong Kong Polytechnic University Hand Dorsal (HD)\footnote{\url{http://www4.comp.polyu.edu.hk/~csajaykr/knuckleV2.htm}} dataset~\cite{KumXu16}. The 11k hands dataset has 11k images with 190 subjects (identities). It also has metadata such as the subject ID, gender, age, skin colour, and a set of information of the captured hand, i.e., right- or left-hand, hand side (dorsal or palmar), and logical indicators referring to whether the hand image contains accessories, nail polish, or irregularities. We exclude any hand image containing accessories from the training data to avoid any potential bias as in~\cite{Mah19}. Unlike the work in~\cite{Mah19} which treats both right and left dorsal or palmar hand of a subject as the same, which is not usually the case and is even not possible to compare to, we treat a subject to be identified by either right or left dorsal, or right or left palmar for robust identification, rather than mixing them since right and left hands show disparity in their vein patterns~\cite{WanCaoTan19}. Accordingly, we separate the dataset into right dorsal, left dorsal, right palmar and left palmar sub-datasets to train a person identification model. After excluding accessories and separating the dataset,  right dorsal has 143 identities, left dorsal has 146, right palmar has 143 and left palmar has 151 identities. The first half and the second half based on the order of identity of each sub-dataset are used for training and testing, respectively. For instance, for right dorsal, the first 72 identities are used for training and the last 71 identities are used for testing. Similarly, the first 73, 72 and 76 identities are used for training phase for left dorsal, right palmar and left palmar, respectively. The remaining identities of each sub-dataset (73 for left dorsal, 71 for right palmar, 75 for left palmar) are used for testing. From each identity of the test set of each sub-dataset, we randomly choose one image and put in a common gallery for all the sub-datasets. The remaining images of each identity of the test set of each sub-dataset are used as a query set for that sub-dataset. Accordingly, the gallery has 290 images and the query has 971 images for right dorsal, 988 images for left dorsal, 917 images for right palmar and 948 images for left palmar. The HD dataset has 502 subjects (identities). The first half and the second half of the dataset based on identity are used for training and testing, respectively i.e. identities of 1 - 251 for training and 252 - 502 for testing. From each identity of the test set, one image is randomly chosen and put in a gallery and the rest are used as a query. There are also additional images of 213 subjects in the HD dataset which lack clarity or do not have second minor knuckle patterns. These additional images of 213 subjects are added to the gallery. Accordingly, the gallery for the HD dataset has 1593 images and the query has 1992 images. A randomly chosen image of each identity of the training set of both datasets is used as a validation for monitoring the training process. This procedure is repeated for 10 times and the average performance is reported. Typical exemplar image (query) of each (sub-)dataset is shown in Fig.~\ref{fig:result} with ranked results retrieved from a gallery.
 
\noindent \textbf{Implementation details:} The GPA-Net is implemented using PyTorch deep learning framework and trained on NVIDIA GeForce RTX 2080 Ti GPU. We train the model using Cross-Entropy loss and mini-batch Stochastic Gradient Descent (SGD) optimizer with momentum; the Nesterov momentum is also enabled. The mini-batch size, the momentum and the weight decay factor for L2 regulization are set to 20, 0.9 and $5 \times 10^{-4}$, respectively. The model is trained for 60 epochs with an initial learning rate of 0.02. We divide this initial learning rate by 10 for the existing layers of the backbone network i.e. bigger initial learning rate is given to the newly added layers (FC layers and batch normalization) with appropriate weight and bias initializations. We also use the learning rate scheduler with a decay factor of 0.1 after every 30 epochs. The input images are resized to $384 \times 384$ and then augmented by random horizontal flip, normalization and color jittering. A random order of images are used by reshuffling the data set. During testing, only normalization is used.

\noindent \textbf{Evaluation metrics:} We use Cumulative Matching Characteristics (CMC)~\cite{ZheSheTia15} (Rank-1 matching accuracy) and mean Average Precision (mAP)~\cite{ZheSheTia15} to evaluate the proposed identification method.

\subsection{Performance Evaluation}

In this section, we evaluate our model on 11k~\cite{Mah19} and HD~\cite{KumXu16} datasets, and we report the results using rank-1 accuracy and mAP. We use input image size of $384 \times 384$ which means the spatial size of \textbf{T} is $24 \times 24$.  After investigating both horizontal and vertical partitions, its ablation analysis is given Table~\ref{tbl:AblationAnalysis}. We find out that partitioning the tensor $T$ into 3 horizontally gives the best performance. In addition to same-domain performance, we also report cross-domain performance in order to prevent being misled by over-fitting. Since there is not much work on person identification based on the full hand, we compare with the baselines ResNet50~\cite{HeZhaSun16} and VGG with batch normalization (VGG-bn)~\cite{SimZisBen15} pretrained on ImageNet by changing the number of classes from 1000 (ImageNet) to the number classes (identities) in our datasets for transfer learning. The training hyperparameter values such as learning rate, batch size, etc. are kept the same for all methods. We use the 2048-dimensional features after the GAP layer of the ResNet50 and the 4096-dimensional output of fc7 of the VGG19-bn as visual representations to compare to our proposed method.

\begin{figure}[htb]
\begin{minipage}[b]{1.0\linewidth}
  \centering
  \centerline{\includegraphics[width=8.5cm]{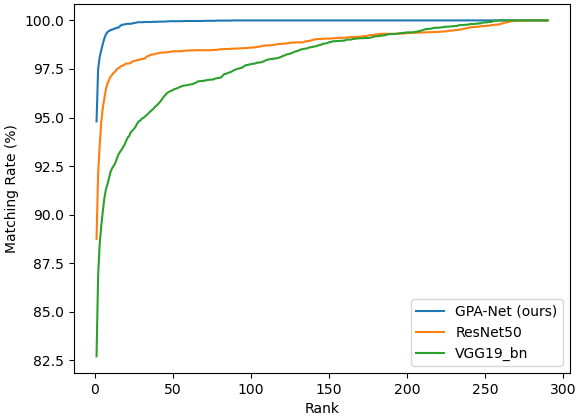}}
\end{minipage}
\caption{Matching rate vs rank comparison of our method with other methods using CMC plot on right dorsal of 11k dataset.}
\label{fig:MatchRank}
\end{figure}

\noindent \textbf{Same-domain performance:} In this setting, the model is trained on a specific dataset and then tested on the test set of that dataset. As shown in Table~\ref{tbl:sameDomain}, our method (GPA-Net) outperforms the other methods by a large margin in both rank-1 accuracy and mAP on all datasets. For instance, it outperforms the baseline ResNe50 by 6.06\% rank-1 accuracy and 5.52\% mAP and the baseline VGG-bn by 12.10\% rank-1 accuracy and 11.27\% mAP on right dorsal (D-r) sub-dataset of 11k dataset. Matching rate versus rank comparison of our method with other methods using CMC plot on right dorsal of 11k dataset is shown in Fig.~\ref{fig:MatchRank}.

\begin{table*} [htbp]
\begin{center}
  \begin{tabular}{|l|ll|ll|ll|ll|ll|}
    \hline
    \multirow{2}{*}{Method} & 
      \multicolumn{2}{c|}{D-r of 11k} & 
      \multicolumn{2}{c|}{D-l of 11k} & 
      \multicolumn{2}{c|}{P-r of 11k} &
      \multicolumn{2}{c|}{P-l of 11k} &
      \multicolumn{2}{c|}{HD} \\
      \cline{2-11}
    & rank-1 & mAP & rank-1 & mAP & rank-1 & mAP & rank-1 & mAP & rank-1 & mAP \\
    \hline
    VGG19-bn & 82.70 & 84.45 & 88.64 & 89.82 & 85.09 & 86.88 & 89.98 & 91.31 & 86.90 & 88.04 \\ 
    ResNet50 & 88.74 & 90.20 & 91.08 & 92.12 & 91.09 & 92.18 & 93.30 & 94.02 & 89.67 & 90.63 \\ 
    \textbf{GPA-Net (ours)} & \textbf{94.80} & \textbf{95.72} & \textbf{94.87} & \textbf{95.93} & \textbf{95.83} & \textbf{96.31} & \textbf{95.72} & \textbf{96.20} & \textbf{94.64} & \textbf{95.08} \\
    \hline
  \end{tabular}
\end{center}
\caption{Same-domain performance comparison of our method with other methods on right dorsal (D-r) of 11k, left dorsal (D-l) of 11k, right palmar (P-r) of 11k, left palmar (P-l) of 11k and HD datasets. Rank-1 accuracy (\%) and mAP (\%) are shown.}
\label{tbl:sameDomain}
\end{table*}
\noindent

\noindent \textbf{Cross-domain performance:} In this case, the model is trained on the training set of one dataset and then evaluated on the test set of the other dataset. We conduct a cross-domain performance on the right dorsal (D-r) of 11k and HD datasets as shown in Table~\ref{tbl:crossDomain}. D-r (11k) $\rightarrow$ HD shows the model is trained on the training set of D-r (11k) dataset and then evaluated on the test set of HD dataset. Our proposed method outperforms the other methods by a large margin on both datasets which shows our method has more generalizability. For instance, our method outperforms the baseline ResNet50 by 24.74\% rank-1 accuracy and 22.57\% mAP and the baseline VGG-bn by 40.42\% rank-1 accuracy and 37.82\% mAP on the D-r (11k) $\rightarrow$ HD.

\begin{table}[htbp]
\begin{center}
  \begin{tabular}{|l|ll|ll|}
    \hline
    \multirow{2}{*}{Method} & 
      \multicolumn{2}{c|}{D-r (11k) $\rightarrow$ HD} & 
      \multicolumn{2}{c|}{HD $\rightarrow$ D-r (11k)} \\
      \cline{2-5}
 & rank-1 & mAP & rank-1 & mAP  \\
\hline
VGG19-bn & 42.60 & 46.83 & 58.87 & 62.24  \\ 
ResNet50 & 58.28 & 62.08 & 69.18 & 72.68   \\ 
\textbf{GPA-Net (ours)} & \textbf{83.02} & \textbf{84.65} & \textbf{86.84} & \textbf{89.06} \\
\hline 
\end{tabular} 
\end{center}
\caption{Cross-domain performance comparison on right dorsal (D-r) of 11k and HD datasets. Rank-1 accuracy (\%) and mAP (\%) are shown.}
\label{tbl:crossDomain}
\end{table}
\noindent

\subsection{Ablation Study}

As described in Section~\ref{sec:proposedMethod}, our method considers both horizontal and vertical partitions on the 3D tensor \textbf{T} of activations in addition to the global 3D tensor. We make an ablation analysis of our method on both partitions and components as follows.

\noindent \textbf{Ablation analysis on partitions:} In this case, we investigate how partitioning in horizontal and vertical directions affects the performance of our method. As shown in Table~\ref{tbl:AblationAnalysis}, we increase the horizontal partitions ($H_p$) and the vertical partitions ($V_p$) from 1 to 4 and from 1 to 3, respectively. The values at $H_p = 1$ and $Vp = 1$  (rank-1 accuracy of 91.19\% and mAP of 92.14\%) show the global compenent of our method (only 1 classifier is trained). Similarly, the values at $H_p = 2$ and $Vp = 2$ show the GPA-Net performance with 1 global classifier and 4 local classifiers. Accordingly, horizontal partitioning with $H_p = 3$ (1 global and 3 local classifiers) gives the best performance with 95.83\% rank-1 accuracy and 96.31\% mAP.

\begin{table}[htbp]
\begin{center}
\begin{tabular}{|c|c|c|c|}
\hline
$H_p$ $\backslash$ $V_p$ & 1  & 2 & 3 \\ 
\hline
1 & 91.19 (92.14) & 93.45 (94.40) & 92.16 (93.20) \\ 
\hline 
2 & 93.05 (94.01) & 94.06 (94.83) & 93.60 (94.21)   \\ 
\hline 
3 & \textbf{95.83 (96.31)} & 95.39 (95.95) & 94.07 (94.80) \\
\hline
4 & 95.42 (95.95) & 94.41 (95.08) & 93.81 (94.50) \\
\hline 
\end{tabular} 
\end{center}
\caption{Ablation analysis of our method across both horizontal ($H_p$) and vertical ($V_p$) partitioning on right palmar (P-r) of 11k hands dataset. The results are shown in \textit{rank-1 (mAP)} format in \%.}
\label{tbl:AblationAnalysis}
\end{table}
\noindent


\noindent \textbf{Ablation analysis on components:} Our proposed method (GPA-Net) has global and local components which learn global and part-level features, respectively, for robust discriminability. 
The learned part-level features outperforms the learned global features. Our method which integrates these components outperforms the components when treated separately, in both evaluation metrics, as shown in Table~\ref{tbl:ComponentAblation}. When evaluated on the right palmar (P-r) of 11k hands dataset, for instance, using only the global features gives 91.19\% rank-1 accuracy and 92.14\% mAP, using only the local features gives 94.33\% rank-1 accuracy and 94.91\% mAP, and using the proposed GPA-Net gives 95.83\% rank-1 accuracy and 96.31\% mAP.

\begin{table}[htbp]
\begin{center}
\begin{tabular}{|c|cc|}
\hline
Method & rank-1 (\%) & mAP (\%) \\ 
\hline
Global only (ours) & 91.19 & 92.14  \\ 
Local only (ours) & 94.33 &  94.91   \\ 
\textbf{GPA-Net (ours)} & \textbf{95.83} & \textbf{96.31} \\
\hline 
\end{tabular} 
\end{center}
\caption{Component analysis of our method on right palmar (P-r) of 11k hands dataset using rank-1 (\%) and mAP (\%).}
\label{tbl:ComponentAblation}
\end{table}
\noindent

%

\section{Conclusion} \label{sec:Conclusion}

In this paper, we introduce a Global and Part-Aware Network (GPA-Net) to learn both global and local (part-level) deep feature representations for person identification based on the hand. The combination of global and part-level feature representations allows a deeper study of the features of the hand in less controlled situations and our extended uniform partitioning helps to obviate the need for explicit pose estimation that can result in compound error. The experimental results on two public hand datasets demonstrate the superiority of the proposed method over the competing methods, giving it strong potential for robust identification of the perpetrators of serious crime.

\bibliographystyle{IEEEbib}
\bibliography{refs}

\end{document}